\documentclass[journal]{IEEEtran} 
\usepackage{amsmath,amsfonts} 
\usepackage{algorithmic}
\usepackage{algorithm}
\usepackage{array}
\usepackage[caption=false,font=normalsize,labelfont=sf,textfont=sf]{subfig}
\usepackage{textcomp}
\usepackage{stfloats}
\usepackage[hyphens]{url}
\usepackage{verbatim}
\usepackage{graphicx}
\usepackage{cite}

\usepackage[numbers]{natbib}
\usepackage{xcolor}
\usepackage{tablefootnote}

\begin{document}

\title{Improving Conditional Level Generation using Automated Validation in Match-3 Games}

\author{Monica Villanueva Aylagas, Joakim Bergdahl, Jonas Gillberg, Alessandro Sestini, Theodor Tolstoy, Linus Gissl\'en 
\thanks{M. Villanueva Aylagas, J. Bergdahl, A. Sestini and L. Gissl\'en are with SEED - Electronic Arts (EA)\\
J. Gillberg was with Electronic Arts (EA) while conducting this research\\
T. Tolstoy is with TrackTwenty - Electronic Arts (EA)}
}


\markboth{\tiny This is the author's version of the article published in IEEE Transactions on Games. Citation information: DOI 10.1109/TG.2024.3440214}
{M. Villanueva Aylagas
\MakeLowercase{\textit{(et al.)}:
Improving Match3 Level Generation with Validation}} 

\IEEEpubid{\begin{tabular}[t]{@{}l@{}}\copyright~2024 IEEE. Personal use is permitted, but republication/redistribution requires IEEE permission.\\See https://www.ieee.org/publications/rights/index.html for more information\end{tabular}}

\maketitle

\begin{abstract}

Generative models for level generation have shown great potential in game production. However, they often provide limited control over the generation, and the validity of the generated levels is unreliable.
Despite this fact, only a few approaches that learn from existing data provide the users with ways of controlling the generation, simultaneously addressing the generation of unsolvable levels. 
This paper proposes Avalon, a novel method to improve models that learn from existing level designs using difficulty statistics extracted from gameplay. 
In particular, we use a conditional variational autoencoder to generate layouts for match-3 levels, conditioning the model on pre-collected statistics such as game mechanics like difficulty and relevant visual features like size and symmetry. Our method is general enough that multiple approaches could potentially be used to generate these statistics. We quantitatively evaluate our approach by comparing it to an ablated model without difficulty conditioning. Additionally, we analyze both quantitatively and qualitatively whether the style of the dataset is preserved in the generated levels. 
Our approach generates more valid levels than the same method without difficulty conditioning.
\end{abstract}


\section{Introduction}
\IEEEPARstart{P}{rocedural} Content Generation (PCG) has been widely used in video game development to increase replayability, present the player with personalized content, or ease the burden of intensive live service content creation, such as in mobile games. In the past, PCG strategies mainly included rule-based generation or planning. The introduction of Procedural Content Generation through Machine Learning (PCGML) \cite{survery_pcgml} has the potential to revolutionize the field by reducing the need for hand-crafted elements, using techniques that take advantage of existing content or learn through interaction with the environment.


Most works that learn from existing data, including human-generated data, use generative models. However, these models are not infallible and do not always yield the desired results. For example, they tend to commit mistakes that are evident to humans. In the context of level creation, mistakes can mean that the level is unplayable or unsolvable, e.g. important elements like keys or doors are inaccessible or missing. 
Studies focusing exclusively on the generation and disregarding the validation of level playability, such as the work done by \citet{doom_generation} and \citet{vae_blending}, have limited practical applications in game production. A common validation approach is to perform post-generation testing, where various methods such as game-specific playability checks or heuristic game-playing agents can be used \cite{attention_gan_zelda, evolutionary_gan_mario}. However, this approach does not improve the generation process but rather confirms whether the generated levels are valid.
\IEEEpubidadjcol

Other previous works have tried to solve the problem of playability by creating levels with a generator that learns from scratch as a Reinforcement Learning (RL) task \cite{controllable_generators_RL, growing_complexity} or guided by an RL validator in an adversarial fashion \cite{ARLPCG_seed, generate_from_nothing}. These approaches generate playable levels but do not allow learning from examples and thus, fail to capture the style of human designers. 

Even in the same genre, different games can have very different level design styles. However, level style is often fairly uniform over a particular game to maintain unified guidelines for all designers and preserve a cohesive gameplay experience. Similar to art style, it is used to give more nuance to a game and differentiate it from other games in the same genre. For an automated solution to be useful in a production environment, it is essential to pick up the small but noticeable differences to adhere to the game's uniform look and feel throughout the progression of levels. Examples of fine-grained design styles can include local and global patterns, artistic choice or intended gameplay. In some cases, the length of the experience can be limited as part of the design style to ensure consistency. The work by \citet{constrained_gans} is the closest to ours, where logical constraints are embedded into a GAN, improving the generation of valid levels while learning from examples. However, this approach requires prior knowledge and the constraints must be mathematically formalized.


We propose Auto-Validated Level Generation (\textit{Avalon}), a framework that can enhance generative models using difficulty validation information during training to improve level validity. We name this approach auto-validation to emphasize that by guiding the generator to produce valid levels during training, the need for post-generation validation is reduced. Our method learns from examples to capture the style of designers, like patterns in existing levels, while guiding the generation by leveraging bot playthroughs that validate if the level can be completed and how. The validity of a level correlates with its difficulty, where the difficulty of an unplayable level tends towards infinite. We apply this framework to generating layouts for a simplified match-3 game, using the number of moves as a proxy for difficulty. 

Casual mobile gaming is a favorable environment for level generation due to the demand for continued content creation as a live service and the opportunity to offer personalized content to maintain player engagement. In particular, the match-3 genre, a type of tile-matching games, represents $21\%$ of the U.S. iPhone game market's revenue as of 2020 \cite{match3_marketshare} which makes it a suitable use case for our proposed method.

Our main contribution consists of a model with the following features:
\begin{itemize}
    \item Novel conditioning mechanism utilizing difficulty: By conditioning on the difficulty it is possible to improve the validity of the generated levels as shown in our ablation.
    \item Valid and stylized generation: Our generator allows designers to use existing levels to generate new ones with similar patterns while creating more valid levels.
    \item Flexible validation approach: Leveraging statistics related to gameplay reduces the need for domain knowledge of the game. Different sources of validation can potentially be used, e.g. heuristics, RL, human testers.
\end{itemize}



\section{Background and Related Work}


\noindent Procedural Content Generation for Games, PCG-G or PCG for short, is an area of research that studies the use of algorithms for creating game content, including textures, levels, behaviors, and more. Application of PCG can be achieved using a large variety of methods, such as generative grammars, spatial algorithms, or artificial intelligence \cite{survey_pcg-g}. One of the most recent approaches to PCG is to use Machine Learning (ML) models as generators. This branch of PCG is called PCGML, and this paper draws inspiration from several studies in this domain. A benefit of PCGML is that it enables generating data by sampling directly from a model that has been trained on existing game assets. Great sources of taxonomies and state-of-the-art compilations can be found in \citet{survery_pcgml} for PCGML in general and in \citet{survey_pcg_dl} specifically for Deep Learning oriented PCG approaches.

\subsection{Automated Gameplay Validation}
\noindent In our approach, the ability to perform automated gameplay validation is crucial. Using automated validation for gameplay and testing is an approach that has attracted attention for its scalability and efficiency in producing high volumes of quantitative results. Recently, the approach has also been used as a way of moving validation and testing ``upstream'', i.e., performing them in the design phase, allowing for more rapid iteration times without waiting for human feedback. Our approach is agnostic to the type of automated validation system, and the only requirement is that the behavior of the gameplay validator is similar enough to that of real players. 
We draw inspiration from previous work in this area. For example, \citet{sestini2022automated} use RL agents infused both with curiosity and expert data that validates gameplay by exploring the proximity of user-recorded trajectories. 
Player modeling techniques such as Deep Player Behaviour Modeling (DPBM) can also be used, similar to \citet{pfau2020dungeons}.

There are several approaches that use PCGML to generate levels automatically validated. For example, \citet{ARLPCG_seed} use a dual-agent system where a \textit{Generator} agent creates levels for a \textit{Solver} agent to play in; both trained in alternation in an adversarial fashion, resulting in levels that are ultimately playability validated. Similar to the proposed method of this study, an auxiliary input is used to enable control over the generated game parameters, like the level of difficulty. However, this approach does not allow the use of previous levels as training data to learn patterns. In \citet{generate_from_nothing}, an RL agent and a generator are used in an end-to-end differentiable solution which, by design, does not need human-crafted examples. The approach allows for self-supervised learning, but as stated by the authors, all the generated levels are too easy to solve for a human. Similarly, \citet{growing_complexity} propose an approach that does not rely on training data but avoids the problem of reward shaping by using recurrent auto-regressive generative flow networks as a control mechanism. 

The lack of training data could be advantageous in some circumstances but it is opposed in nature to the goal of our use case, where the style of level designers is important.

\subsection{Controllable PCGML for Level Generation}
\noindent 
Another important aspect of our approach is the ability to let the users control the generation. This can be done in several ways.
\citet{doran2010controlled} propose using genetic algorithms to create intelligent agents that generate terrains with user-defined constraints. These controllable constraints stem from a system of limited, designer-assignable tokens consumed by the agents each time they perform generation actions. Further, \citet{guzdial2018explainable} present a method that uses design patterns to control the output of the generating model.
However, as in the previous section, these approaches do not learn from previous data. Methods that allow controllability while learning from existing levels include methods like path of destruction, generative adversarial networks (GANs), conditional variational autoencoders (cVAEs), among others.

Path of destruction is a data augmentation method that has been used to train iterative generators. In \citet{pathofdestruction}, the method is extended to allow designer control for features like the number of enemies using conditional inputs. However, the random destruction actions that are at the core of the data augmentation makes it difficult to achieve certain feature combinations. The iterative nature of the generation, where the condition inputs are $\{-1, 0, +1\}$ might be cumbersome to game designers.

GANs and VAEs have been used in PCGML in similar ways. One of the applications is to generate levels using controllable mechanisms, such as evolutionary search of the latent space \cite{gan_latent_evolution, doom_latent_evolution, lode_vae_evolution} or using a conditional input \cite{doom_generation, attention_gan_zelda, vae_blending}. Conditional networks avoid running evolutionary search at inference time and moves the controllability to the training stage. One benefit of this approach is faster inference when the fitness function is computationally expensive.

\citet{doom_generation} compare two GANs, conditioned and unconditioned, trained on existing levels. They can generate levels similar to human-designed ones, with the conditioned network achieving better results. However, the results of both this work and its continuation \cite{doom_latent_evolution} are evaluated as images and never validated in the game.
The drawback with generative models is that without additional information, the level alone might not be sufficient for capturing the structural or style-related constraints of the designer. One similar approach to ours adds a control mechanism to GANs where constraints are embedded into the model during training in a novel architecture named Constrained Adversarial Networks \cite{constrained_gans}. However, it requires prior knowledge to define these constraints, and mathematical knowledge to formalize them, which is often difficult. Often, game designers do not possess this knowledge, rendering the approach practically infeasible in certain use cases. As mentioned by the authors, their method complements other approaches when there is no prior knowledge about invalid structures, or the constraints are hard to formalize.

\citet{lode_vae_evolution} compare vanilla autoencoders against VAEs and find that the variational version produces more details. \citet{vae_blending} show that conditional variational autoencoders can be used to generate complete dungeon and platformer levels producing fitting patterns. Additionally, they demonstrate that blending between styles and genres is possible. However, as mentioned in their future work section, playability evaluations are not performed.

\subsection{Match-3 Level Generation}

\noindent In this paper, we target match-3 games with the well-established mechanics of \textit{Bejeweled} \cite{bejeweled_game} and \textit{Candy Crush Saga} \cite{candycrush_game}. In these games, colored tiles can be swapped horizontally or vertically to make them disappear when \textit{matched} with adjacent tiles of the same color. In match-3 games, the matching criteria is to align three or more of such tiles. New tiles are randomly added to the board as others disappear until the player achieves a goal or runs out of moves. In a typical match-3 game, the number of levels can reach over ten thousand, and hundreds must be added weekly to keep the players engaged. This leads to a high output, which can lead to repetitive and predictable levels. Unfortunately, the number of PCG works tackling match-3 games is lower than for other game types.


An example of previous work in match-3 level generation is \citet{lilys_garden} which uses two condition vectors to control a GAN that generates levels with adjustable map-shapes and piece distributions for the game \textit{Lily's Garden} \cite{lilysgarden_game}. Even though multiple quantitative metrics validate that the network follows the conditioners, the validity of the levels is only tested for whether they are unstartable or their layouts are corrupt with level elements located outside the map area. 

\citet{candy_crush} present a taxonomy of patterns in training examples for PCGML and models that capture them. In particular, they compare the use of symmetric neighbors in Markov Random Fields (MRF) and training GANs using purely symmetric \textit{Candy Crush Saga} levels to improve the generation of globally symmetric levels. The evaluation consists of a symmetry score comparison and a user study. Unfortunately, this evaluation does not address how well the model learns the training data but rather if it can create symmetric levels. Moreover, the validity of the generated levels is not measured.

Our work differs from the mentioned studies as it aims at generating levels that follow the style of the training set while improving the playability of the generated levels. Additionally, we provide the user with low-level controllability in the generation in terms of the desired difficulty of the level.



\section{Method}
\label{sec:method}

\noindent The previous section shows that most works focus on two main approaches. The first one is using a validator to guide the generation process without prior information or domain knowledge. The second one is using generative models and a dataset of preexisting levels, potentially validating the generation as a post-generation step. 
However, the first approach can converge in levels that are too easy to solve or are not in line with the intent and style of level designers, while the second can commit mistakes that render the level unplayable.

\begin{figure}[!t]
    \centering
    \includegraphics[width=0.48\textwidth]{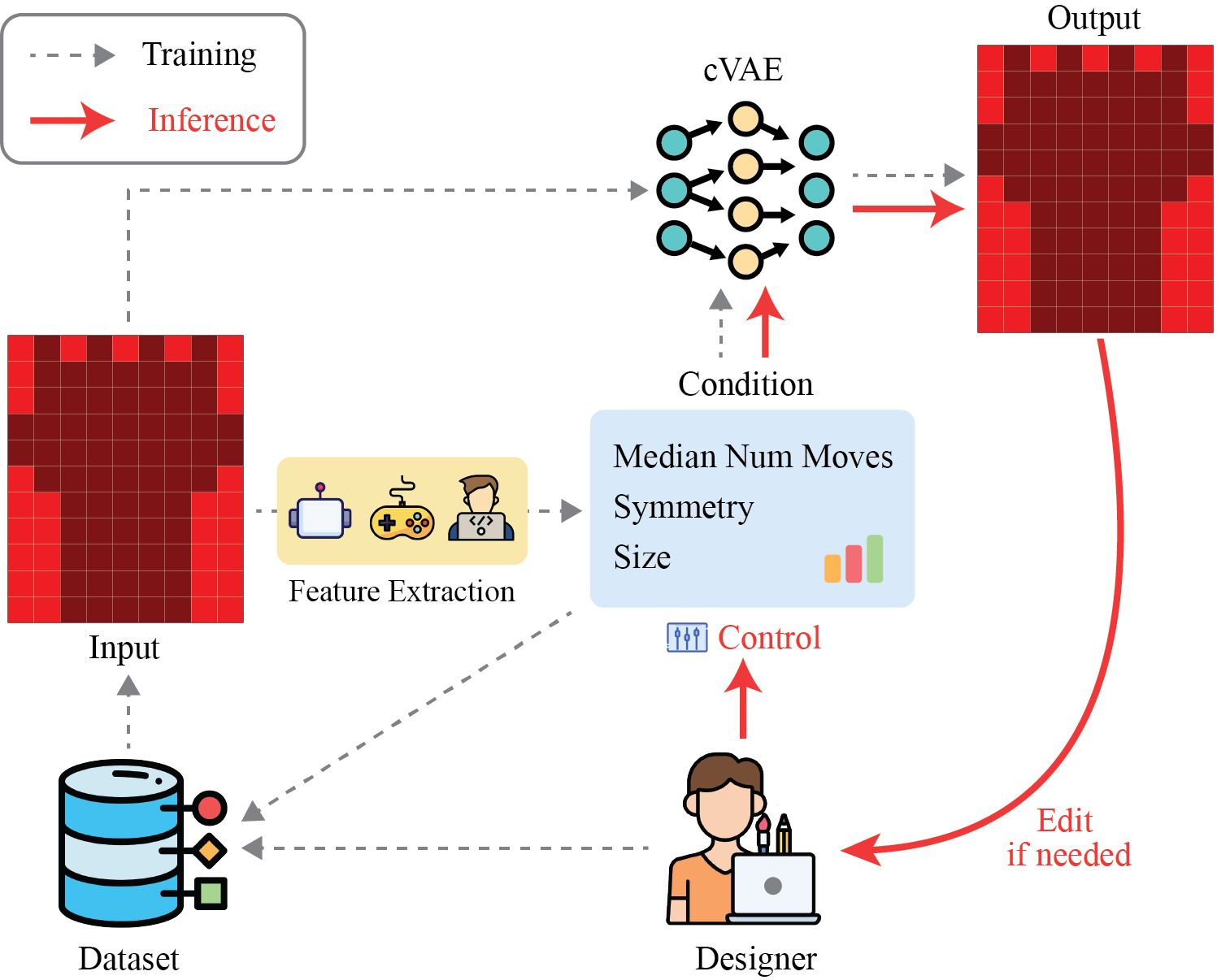}
    \caption{Avalon approach. Before training, -- gray dotted lines in the figure -- the dataset is constructed with levels created by level designers. Offline, we extract a set of features for each level: the median number of moves required to solve it, as well as the board size, type of symmetry. For each level, we use these features as conditioners for the model during training. These features can be extracted using scripted bots or other methods (eg. RL agents or game testers). For our experiments, we use the former approach and train a conditional variational autoencoder. During inference, -- red continuous lines in the figure -- the designer controls the generation through the conditional features, and can manually edit the output level or generate a new one.}
    \label{fig:system_overview}
\end{figure}

We propose a hybrid method called \textit{Avalon}, that uses generative models to learn existing patterns and enhances them through statistics obtained from validating the training examples using game-playing agents.
We apply this framework to a simplified match-3 game to generate layouts for a fixed objective, considering valid levels those that can be solved in the maximum number of moves determined by designers.
In particular, we use a conditional variational autoencoder (cVAE) \cite{vae, cvae} and enhance it with conditioning information based on the number of moves necessary to solve the level. The approach is illustrated in Figure \ref{fig:system_overview}.
Information about the game, level representation, and the dataset can be found in Section \ref{sec:data}. 



\subsection{Condition Design}
\label{sec:condition-design}

\begin{figure}[!t]
    \centering
    \includegraphics[width=0.48\textwidth]{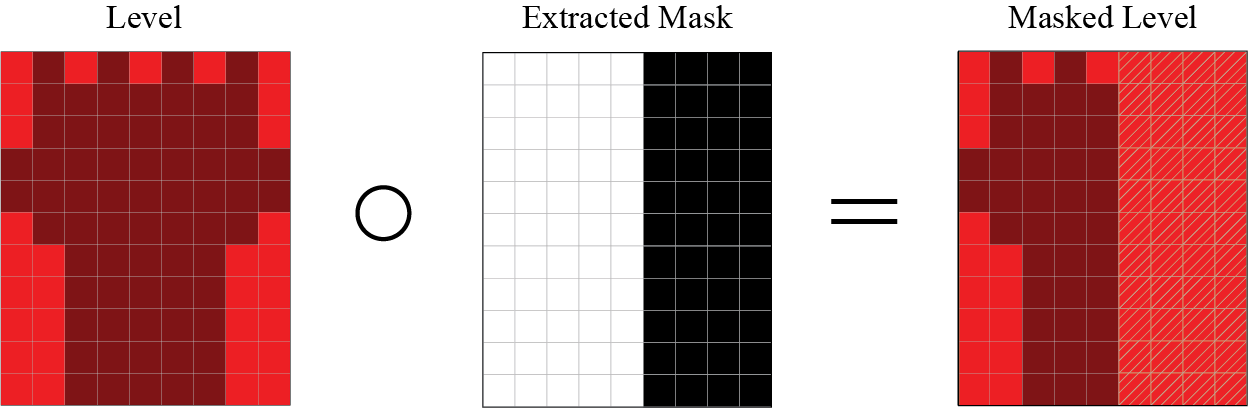}
    \caption{Example application of a vertical symmetry mask. The mask is first extracted from a level in the dataset, and then it is applied to the same level following $L \circ M^{\text{sym}}$. The masked level is part of the training dataset.}
    \label{fig:mask_application}
\end{figure}

\noindent The \textit{Avalon} generator is based on a simple cVAE architecture. For the match-3 level generation implementation, we condition the network on the median number of moves needed to solve the level as a proxy for difficulty and validity. This design decision allows flexibility at inference time when validating if the generated level can be solved in a number of moves different than originally defined by the designers. In our implementation, we use a scripted bot to play the game offline and compute and save the number of moves in the dataset. In addition, we also decided to use conditioners to control features that were deemed essential, such as symmetry, as in \citet{candy_crush} and size, similar to \citet{lilys_garden}.

To avoid the symmetry and global pattern generation problem discussed in \citet{candy_crush}, we developed a novel partial generation cVAE based on masking. Empirical experimentation indicates that this approach works much better than the traditional generation, where the symmetry accuracy peaked around $15\%$. 

To adapt the cVAE framework to partial generation, we preprocess the input data according to the type of symmetry present in the sample using masks. The types of symmetry we consider in this work are: vertical, horizontal, quadrant (vertical and horizontal) and unknown. 
During the creation of the dataset, a level sample $L$ is analyzed and, according to the type of symmetry observed, a mask $M^{\text{sym}} \in \{0,1\}^{W \times H}$ is created and coupled to it. $W$ and $H$ correspond to the width and height of the level. 

Before encoding, a training level $L$ is masked using its symmetry mask $M^{\text{sym}}$ following $L \circ M^{\text{sym}}$, the element-wise product. This masking ensures there is consistency over levels with the same symmetry in the latent space. The symmetry masks declare which layout cells should be used when training, represented by $1$, and which cells should be ignored, represented by $0$. For vertical and horizontal symmetries, the non-zero elements of the mask are designed to cover the available play area up until and including its center point width- and height-wise respectively. For quadrant symmetry, both of these conditions apply. An example of the mask application for vertical symmetry is illustrated in Figure~\ref{fig:mask_application}.
As we will see in Section~\ref{sec:training}, when computing the reconstruction loss, this mask is used to ignore the redundant part of the level. 

During inference, the output is postprocessed according to the symmetry input, duplicating the relevant part of the level if necessary, ensuring that the symmetry requirement is met. 
Similar to the symmetry mask, a size mask $M^{\text{size}}$ is also created for each level in the dataset. To allow for any input level shape, a fixed board width and height are set based on the largest playable area defined by level design. For smaller levels, the playable area is placed at the center and the size mask $M^{\text{size}}$ is used to exclude cells beyond the playable area up until the maximum width and height, encoded with $0$.

\subsection{Architecture}
\label{sec:architecture}
\noindent The encoder is a Neural Network (NN) with three convolutional layers of $16,32$ and $64$ filters and $3\times3$ kernels followed by ReLU activations. The distribution statistics are the output of two independent fully connected (FC) layers that downsample the output 
from the convolutions into a $5$-dimensional space.

The decoder receives $\hat{\mathbf{z}}$, a latent vector $\mathbf{z}$ concatenated with the condition information. The decoder NN is composed of an FC layer that upsamples this input to $64$ dimensions, followed by three layers of transposed convolutions of $32,16$ and $K$ filters, where $K$ is the number of categorical values, and $3\times3$ kernels mirroring the encoder. Each layer is followed by a ReLU activation except for the last one.

The conditional information is used in the encoder and decoder in different formats.
For the encoder, the size conditioner is a mask $M^{\text{size}}$ that defines the play area.
The difficulty conditioner $d$ is the number of moves normalized to the $[0, 1]$ range according to the values observed in the training set. By repeating this value in a 2D map $D \in \mathbb{R}^{W \times H}$, it can be concatenated with the size mask $M^{\text{size}}$ and the symmetry masked level to produce the final input to the encoder $\hat{Y}_{n} = [L_{n} \circ M^{\text{sym}}_{n}, M^{\text{size}}_{n}, D_{n}]$ for a given level $L_{n}$ in the dataset, where $M^{\text{sym}}_{n}$ and $M^{\text{size}}_{n}$ are its respective symmetry and size masks. Since the input data is preprocessed and symmetry is enforced as a postprocessing step, using symmetry as part of the conditioning information is unnecessary.
In the decoder, the size conditioner is in the form of two one-hot vectors, $\mathbf{h}_{W}$ for the width and $\mathbf{h}_{H}$ for the height. These conditioners are concatenated to the sampled latent vector to create the input to the decoder $\hat{\mathbf{z}} = [\mathbf{z}, \mathbf{h}_{W}, \mathbf{h}_{H}, d]$, with $d$ being the aforementioned normalized number of moves.

Experiments with different types of network architectures and conditionings  were performed, and the best experimental results according to the inference metrics are reported here. For transparency, some of the architectural experiments performed included changing the number of layers and the kernel size, using a hierarchical approach, and using different activation functions (eg. ReLu, leaky ReLu). On the conditional side, the experiments tested the use of absolute and normalized values for the difficulty. Another experiment studied the encoding of the size in a similar way to the difficulty, that is, as a categorical or one-hot encoding repeated over a $2D$ map. Additionally, we tested encoding the size information both as separate $W$ and $H$ and as a holistic measurement.

\subsection{Training}
\label{sec:training}
\noindent Except for the modifications needed for the partial generation approach, the network follows the standard VAE loss in Equation \ref{eq:final_loss}. We define our data as a categorical distribution, thus using cross-entropy loss (Equation \ref{eq:cross-entropy_loss}) as the reconstruction loss.

\begin{align}
    \mathcal{L}_{CE}(\hat{X}, \hat{\mathbf{z}}) &= \mathbb{E}_{q_{\phi}(\mathbf{z}|\hat{X})}[\log p_\theta(\hat{X}|\hat{\mathbf{z}})] \label{eq:cross-entropy_loss}\\
    \mathcal{L}_{KL}(\mathbf{z},\hat{Y}) &= - D_{KL}(q_\theta(\mathbf{z}|\hat{Y}) || \mathcal{N}(\mathbf{0},\mathbf{I})) \\
    \mathcal{L} &=\mathcal{L}_{CE} + \mathcal{L}_{KL} \label{eq:final_loss}
\end{align}

where $\hat{X}$ is the output logits of the decoder masked with the symmetry mask, following $\hat{X} = X \circ M^{\text{sym}}$; $\hat{Y} = [L \circ M^{\text{sym}}, M^{\text{size}}, D]$ is the encoder input defined as the symmetry masking $M^{\text{sym}}$ applied to ground truth level $L$ concatenated to the conditioner as described in Section~\ref{sec:condition-design}; $\hat{\mathbf{z}} = [\mathbf{z}, \mathbf{h}_{W}, \mathbf{h}_{H}, d]$ is the input to the decoder, defined as the concatenation of the sampled latent vector with the conditioners as explained in Section~\ref{sec:architecture}. The ground truth level is represented as $L \in \{0,..,K\}^{W \times H}$ where $K$ corresponds to the number of categorical tile classes.

We train the network for $24000$ epochs using batches of $100$ samples and taking checkpoints every $500$ epochs. The optimizer used is Adam \cite{adam} with a learning rate of $10^{-5}$ for \textit{Avalon} and $5\times 10^{-6}$ for \textit{Vanilla}.


\section{Experiments and Evaluation}
\label{sec:eval}




\noindent This section introduces the match-3 formulation we use in our work and the motivation for the datasets, models and evaluation framework employed. \\

\begin{figure*}[!t]
\centering
\subfloat[]{\includegraphics[width=3.in]{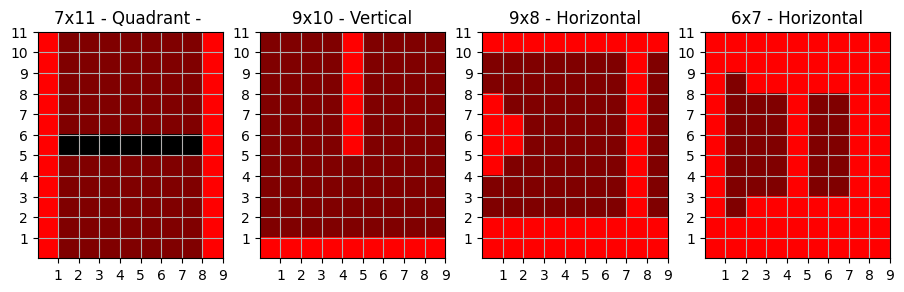}%
\label{fig:dataset_main_example}}
\hfil
\subfloat[]{\includegraphics[width=3.in]{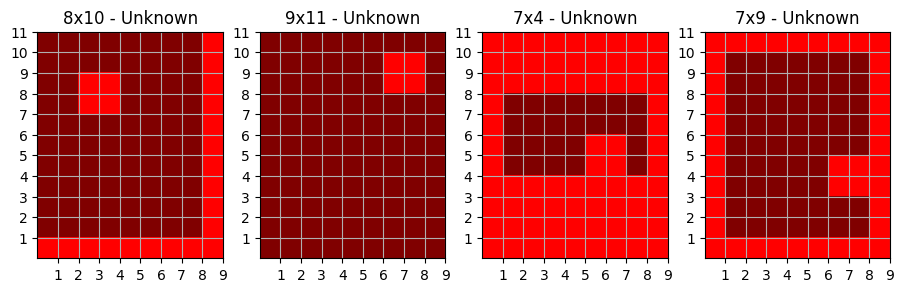}%
\label{fig:dataset_stylized_example}}
\hfil
\subfloat[]{\includegraphics[width=3.in]{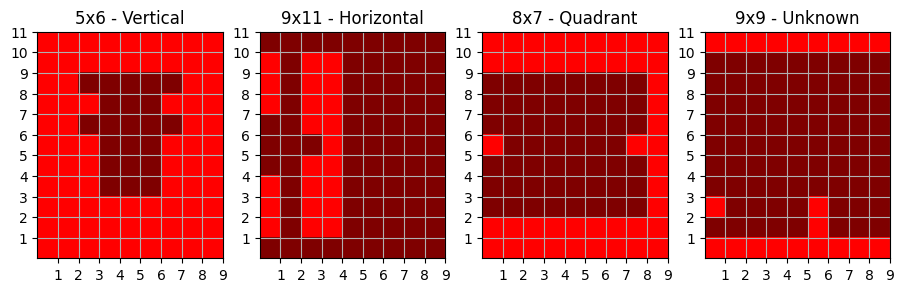}%
\label{fig:inference_vanilla_example}}
\hfil
\subfloat[]{\includegraphics[width=3.in]{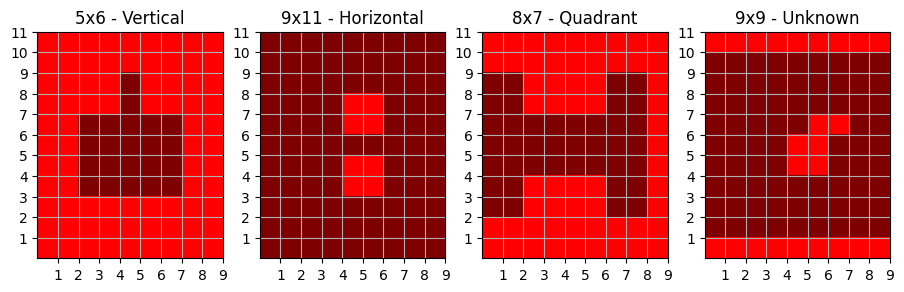}%
\label{fig:inference_vanillastylized_example}}
\hfil
\subfloat[]{\includegraphics[width=3.in]{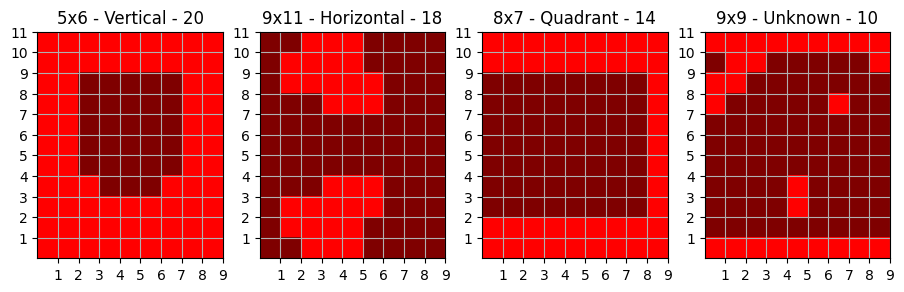}%
\label{fig:inference_avalon_example}}
\hfil
\subfloat[]{\includegraphics[width=3.in]{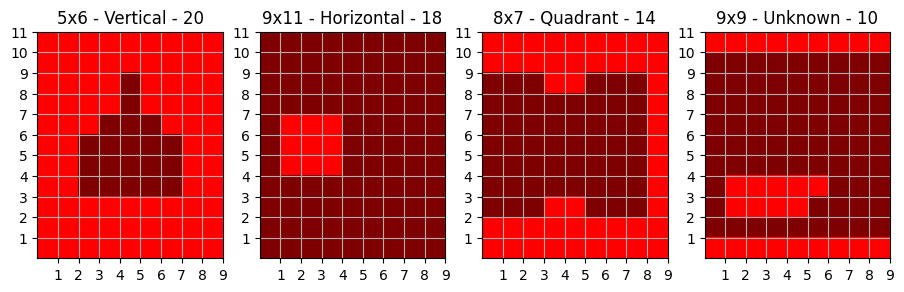}%
\label{fig:inference_avalonstylized_example}}
\caption{Image representation of the levels. The black pixels represent GAP cells, the red pixels depict BLOCK cells and the dark red pixels correspond to PLAYFIELD cells. (a) Training examples from the main dataset (b) Training examples from the stylized dataset (c) Inference examples from the \textit{Vanilla} generator. (d) Inference examples from the \textit{VanillaStylized} generator. (e) Inference examples from the \textit{Avalon} generator. (f) Inference examples from the \textit{AvalonStylized} generator. The text in subplots (c) through (f) indicates the conditioners used: for sublopts (c) and (d), the text indicates the size (e.g. 5x6) and the symmetry (e.g. Vertical) used as input to \textit{Vanilla} and \textit{VanillaStylized}, while subplots (e) and (f) include also the number of moves (e.g. 20) used as input to \textit{Avalon} and \textit{AvalonStylized}.}
\label{fig:examples}
\end{figure*}

\subsection{Baselines}
\noindent Our method aims to generate levels that follow the style of the dataset while reducing the computational cost of generating invalid levels that need to be regenerated. In match-3, almost any layout is solvable given enough moves. However, the players are restricted to do so in a predefined number of moves, which in our case is limited by the level designers' guidelines to 20 moves.

Thus, our evaluation focuses on three main areas: 1) Analyzing the validity of the levels generated through our approach, 2) Validating whether the network we have defined can be trained to accurately capture and generate levels in a style consistent with that of the dataset, and 3) Assessing the network's accuracy in following the level designers' input. 

We compare the following approaches:
\begin{itemize}
    \item \textit{Avalon}: Our novel framework defined in Section~\ref{sec:method}.
    \item \textit{Vanilla}: An ablation to the \textit{Avalon} approach. It uses the same training setup and neural network architecture, including the size and symmetry conditioners but \emph{without difficulty condition}. This is similar to other approaches in the literature that do not condition the generator with game statistics like difficulty.
    \item \textit{AvalonStylized}: Follows the same approach as \textit{Avalon} but is trained using only a dataset of examples with a well-defined pattern, used to evaluate 2).
    \item \textit{VanillaStylized}: Ablation to the \textit{AvalonStylized} approach. As in \textit{Vanilla}, it uses the same setup of \textit{AvalonStylized} but without difficulty condition.
    
\end{itemize}


\subsection{Data}
\label{sec:data}

\noindent In our simplified match-3 game, the level layout allows for boards of sizes between $4\times4$ and $9\times11$ $W\times H$. These layouts are made up of three types of cells: PLAYFIELD cells define locations in the grid where tiles can be placed, and the game can be played; BLOCK cells define the size and shape of the level to which tiles are constrained; and GAP cells resemble BLOCK cells visually, but allow tiles to fall through. The types of symmetry considered are vertical, horizontal, quadrant (vertical and horizontal), and unknown.
The game is simplified and assumes the goal of the level is to match $60$ red pieces in a maximum of $20$ moves, where the color pieces (red, green, blue and orange) are spawned from certain spots by sampling from a uniform distribution with replacement.


\subsubsection{Level Representation}

\noindent To be able to train the cVAE, we convert the level representation into a fixed-sized matrix  $W\times H =9\times11$, where the three types of cells are represented with categorical values, such that a level is defined as $L \in \{0,1,2\}^{9 \times 11}$. The size of the layout is defined as the maximum number of consecutive PLAYFIELD cells in all rows for the width and all columns for the height, with the play area always located in the center (see Figure \ref{fig:examples} for reference). In the decoder, the condition information that forms the input $\hat{\mathbf{z}}$ is represented by $\mathbf{h}_{W}$, a $6$-dimensional one-hot vector encoding the width values, ranging from $[4,9]$; $\mathbf{h}_{H}$, a $8$-dimensional one-hot for the height values, between $[4,11]$; and a $1$-dimensional float for $d$, the number of moves.

Tile-spawning spots are automatically created as a postprocessing step when converting the model's output into the level representation consumed by the game. Thus, the ML data representation includes neither these spawning spots nor the color pieces they produce. 

\subsubsection{Datasets}
\label{sec:datasets}

\noindent We use two datasets in our evaluation framework. The main dataset is used for all experiments except for the style comparison, where we use the stylized one.

\begin{itemize}
    \item Main dataset: This dataset is composed of $81$ designer levels and $117$ levels created with a rule-based approach. These levels use mainly BLOCK cells, but GAP cells are used to a lower extent. The training data contains $170$ levels, and the validation and test sets contain $15$ and $13$ levels, respectively.
    \item Stylized dataset: This dataset is designed to have a local pattern, that represents a style, easily observed visually. In this case, we have used a single $2\times2$ BLOCK pattern (no symmetries) randomly placed in any position of the board. The training, validation, and test dataset sizes are identical to the main dataset, all procedurally created.
\end{itemize}

\begin{figure}[!t]
    \centering
    \includegraphics[width=0.48\textwidth]{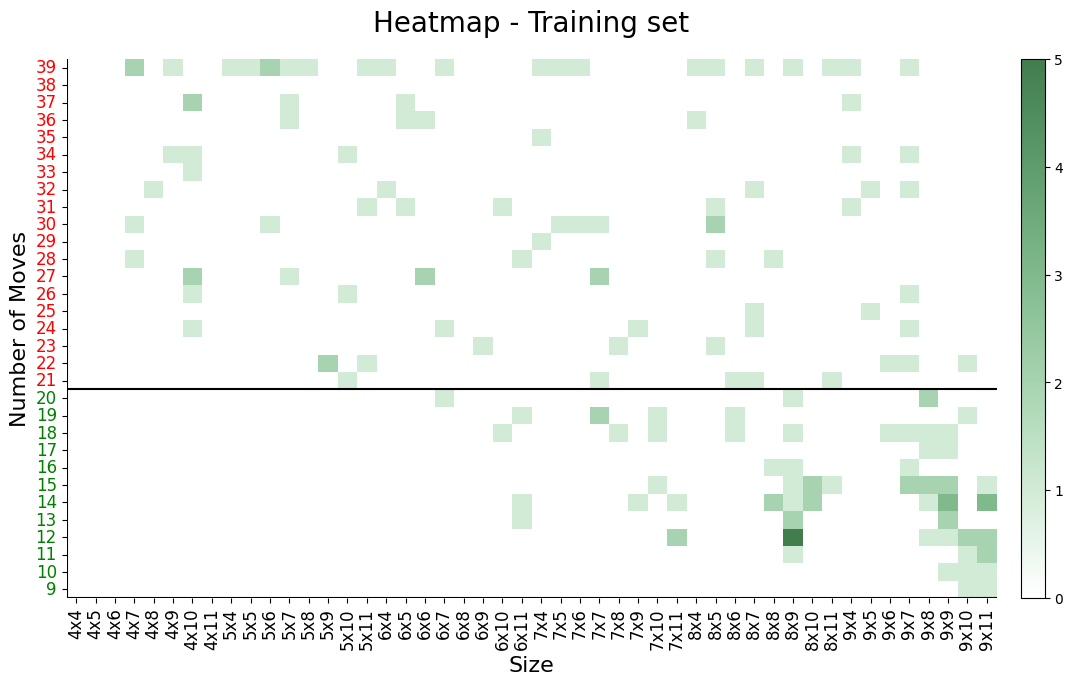}
    \caption{Analysis of the main dataset's training set. The heatmap represents the number of levels for each size (x-axis) and each median number of moves for the level to be solved (y-axis), according to the statistics extracted by our bot. The horizontal line indicates the threshold that separates valid from invalid levels.}
    \label{fig:training_heatmap}
\end{figure}

Each level in the main dataset is evaluated with a scripted bot in $30$ games to ensure convergence, and the median and standard deviation values are stored to be used in training and evaluation. Informal experiments indicate that the bot performs comparatively to human testers in the development team (for reference, a random bot would not solve any of the levels in the training set in less than $20$ moves).
In our framework, the bots are allowed to make $39$ ($2\times \textit{valid\_moves} - 1$) moves, more than the $20$ valid moves, to give designers flexibility in case they want to increase the gameplay length or award additional moves to the players without the need to retrain. This also helps increase the dataset size. Figure \ref{fig:training_heatmap} displays the distribution of training levels in the main dataset regarding their size and number of moves needed for them to be solved. For examples of training samples in both datasets, we refer to Figure \ref{fig:dataset_main_example} and \ref{fig:dataset_stylized_example}.

\subsection{Experimental Setup and Performance}
\label{sec:setup_performance}

\begin{table}[!t]
\caption{Inference times. Results are reported in \emph{ms per level} (non-batched) averaged over $10$ runs. The \textit{Stylized} models are not reported since the architecture is the same as their counterparts.
\label{tab:performance}}
\centering
\begin{tabular}{lll}
\hline
    & \textit{Avalon} (ms/level) & \textit{Vanilla} (ms/level) \\
\hline
{Inference in PyTorch}        & $1.69$          & $0.99$    \\
{Inference in Unity}     & $2.50$          & $2.40$    \\
\hline
\end{tabular}
\end{table}

\noindent All experiments were conducted on a single machine with an NVIDIA RTX A6000 GPU ($48$ GB VRAM), an Intel(R) Xeon(R) Gold 6230 CPU at $2.10$ GHz and $64$ GB of RAM using a training solution based on \textit{PyTorch} \cite{pytorch}.

When generating new levels, we choose the maximum value after a softmax activation. The best model is selected as the checkpoint that maximizes the average of several metrics in inference weighted equally: diversity, size, and tile distribution accuracy (more information in the following sections). 
All the metrics are computed by sampling every possible size ($48$ combinations) using $3$ types of symmetry: vertical, horizontal, and quadrant. This gives a total of $144$ generated levels. For the \textit{Avalon} generator, we sample a difficulty uniformly between the minimum value in the dataset and the maximum number of valid moves, $20$, to avoid a combinatorial explosion and simulate a real scenario. 
Level generation in \textit{Unity} was performed using the \textit{Barracuda} package \cite{barracuda}. Inference times can be examined in Table \ref{tab:performance}. Note that batch inference is faster per level on average.

The simplified match-3 game and the validation bot were implemented in \textit{Unity} \cite{unity}. The same bot is used to compute the statistics used during training and to compute the evaluation metrics.

\subsection{Conditioning Evaluation}

\noindent Level designers can condition the generation on the size of the board, the type of symmetry they want to use, and the median number of moves needed to solve the level for the \textit{Avalon} generator.

Early in development, we noticed that the network could revert to generating empty levels, artificially improving size and symmetry metrics. To prevent this problem, we introduced a diversity metric used in previous works such as \citet{constrained_gans} or \citet{growing_complexity}. Additionally, we report a plagiarism score to evaluate inference novelty.

\begin{itemize}
    \item Size accuracy: is computed as the percentage of levels generated with the exact size used as conditioning. For a single level, the accuracy is $100\%$ if the size is correct and $0\%$ otherwise. This metric is similar to the under/overfilled levelshape metrics in \citet{lilys_garden}.
    \item Diversity accuracy: is defined as the average Hamming distance level-wise between all pairs of levels. For a pair of levels, the accuracy is $100\%$ if two levels are different, even if it is only by one cell, and $0\%$ if the level is duplicated. We use level-wise diversity instead of tile-wise to obtain a more relevant metric due to the general similarity among match-3 levels.
    \item Difficulty accuracy: is a new metric designed as the percentage of generated levels whose difficulty conditioning is within the statistics of the generated level after validation. For a single level, we evaluate it following the same steps as for constructing the dataset (as explained in Section \ref{sec:datasets}). The accuracy will be $100\%$ if the conditional number of moves falls within the standard deviation of the validated level and $0\%$ otherwise. We also compute the mean and standard deviation of the distance between the conditional and the validated number of moves.
    \item Plagiarism score: is computed as the percentage of generated levels that are identical to any other level in the training set. The score is $100\%$ if all the levels generated exist in the dataset and $0\%$ if all the generated levels are original.
    
\end{itemize}

We compare the \textit{Avalon} and \textit{Vanilla} generators trained on the main dataset and report the above metrics for inference. Note that direct comparison with the \textit{Stylized} models is not relevant since they are trained on a different dataset. 

Symmetry accuracy is not reported explicitly since the result of using the partial generation and postprocessing approaches consistently attains $100\%$ accuracy.

\subsection{Playability Evaluation}
\noindent To measure the robustness of our method, we compare the percentage of valid levels generated by the \textit{Avalon} generator as opposed to that of the \textit{Vanilla} generator. 

We define a level as \textit{valid} if it can be solved in the maximum number of moves according to the designers' guidelines, in our case, $20$ moves. We consider levels that can be solved in more than $20$ moves as invalid.

\subsection{Style Evaluation}
\noindent Style evaluation is an open problem for generative models. Previous works such as \citet{attention_gan_zelda} or \citet{lilys_garden} have tried to show that the generation is similar to the dataset by comparing the training and generated distributions of different types of tiles. 

The distribution of tiles can vary substantially depending on the size of the board and, thus, the available playing area. Hence, we have adapted this methodology. Additionally, given the large number of plots to compare and the need for a numerical value to use as part of the framework to select the best model, we have designed a quantitative proxy denominated \textit{tile distribution} accuracy.
We consider the distributions comparable if the median of the inference distribution is contained within the first and the third quartile, $Q_{1}$ and $Q_{3}$, of the training distribution. This metric is a proxy and only partially explains the problem. 

As already mentioned, we train the \textit{VanillaStylized} and \textit{AvalonSylized} generators  on the stylized dataset to show qualitatively how close the results follow a predefined style and how they differ from the results obtained from the models trained on the main dataset.


\section{Results}
\label{sec:results}


\begin{table}[!t]
\caption{Quantitative Results of Two Baseline Models.\\ * The number of valid levels in the dataset is $46.15\%$
\label{tab:results}}
\centering
\begin{tabular}{lll}
\hline
    & \textit{Avalon} ($\%$) & \textit{Vanilla} ($\%$) \\
\hline
{Size accuracy}        & $88.48$         & $91.94$ \\
{Diversity accuracy}     & $96.51$         & $97.93$ \\
Plagiarism score     & $00.00$         & $00.00$ \\
{Difficulty accuracy}       & $33.33$         & N/A  \\
\hline
{Valid levels*}       & $51.39$ & $43.75$ \\
{Tile dist. accuracy}       & $55.81$         & $65.12$   \\
\hline
\end{tabular}
\end{table}

\noindent A summary of the results of the evaluation methodology described in Section \ref{sec:eval} can be found in Table \ref{tab:results}. In the following sections, we will summarize these results by focusing on model comparisons.

\subsection{Avalon vs Vanilla}
\noindent Firstly, we compare our proposed method, \textit{Avalon}, and the ablated alternative without difficulty conditioning, \textit{Vanilla}. Our results show that the \textit{Avalon} model generates more valid levels than the \textit{Vanilla} model, increasing from $43.75\%$ to $51.39\%$. For reference, the dataset contains $78$ ($46.15\%$) valid levels. This indicates that difficulty conditioning does, indeed, help improve the generator in terms of playability. As an interesting note, \textit{Avalon} consistently outperforms \textit{Vanilla} under the same evaluation framework even if the validity threshold is increased to a higher number of moves.

However, adding the difficulty conditioning seems to create a trade-off between the accuracies of the previous learning task (size, diversity, and tile distribution accuracy) and the new requirement. The results in Table \ref{tab:results} show a decrease of 3 percentage points in size accuracy, 1 in diversity accuracy, and 10 in tile distribution accuracy.

The decrease in diversity is not significant compared to the other metrics and can be explained by the solution space reduction introduced by adding a new constraint. However, the size and tile distribution accuracy decrease suggests a potential generation degradation. The tile distribution metric might not be robust enough to describe the problem correctly due to the lack of data for some sizes and the ambiguous correlation between style and distribution (further analyzed in the next section). In the case of size accuracy, and difficulty accuracy, the performance decrease might be related to unreasonable conditional requirements that can result in conflicting characteristics. For example, very small playing areas require more moves to solve the level, as the number of available moves is constrained. Consider a case where a small size is sampled, e.g., $4\times4$, and a low number of moves, e.g. $10$. The generator will be forced to miss the target size by making a bigger level that can be solved in the target number of moves or miss the difficulty target by staying true to the requested size.

Additionally, the low difficulty accuracy together with a significant distance between the conditional and validated number of moves (average distance of $\mu=10.19$ and standard deviation $\sigma=8.93$) seems to point towards other problems, including the sparsity of valid levels in the dataset (see Figure \ref{fig:training_heatmap}) and the lack of a validation metric that can help choose a checkpoint that maximized the difficulty accuracy (presented in Section \ref{sec:setup_performance}).

It is worth noting that the plagiarism score is $0$, which indicates that all of our generated levels are original, i.e. are not contained in the training set.

\subsection{Style Analysis}

\begin{figure}[!t]
    \centering
    \includegraphics[width=0.45\textwidth]{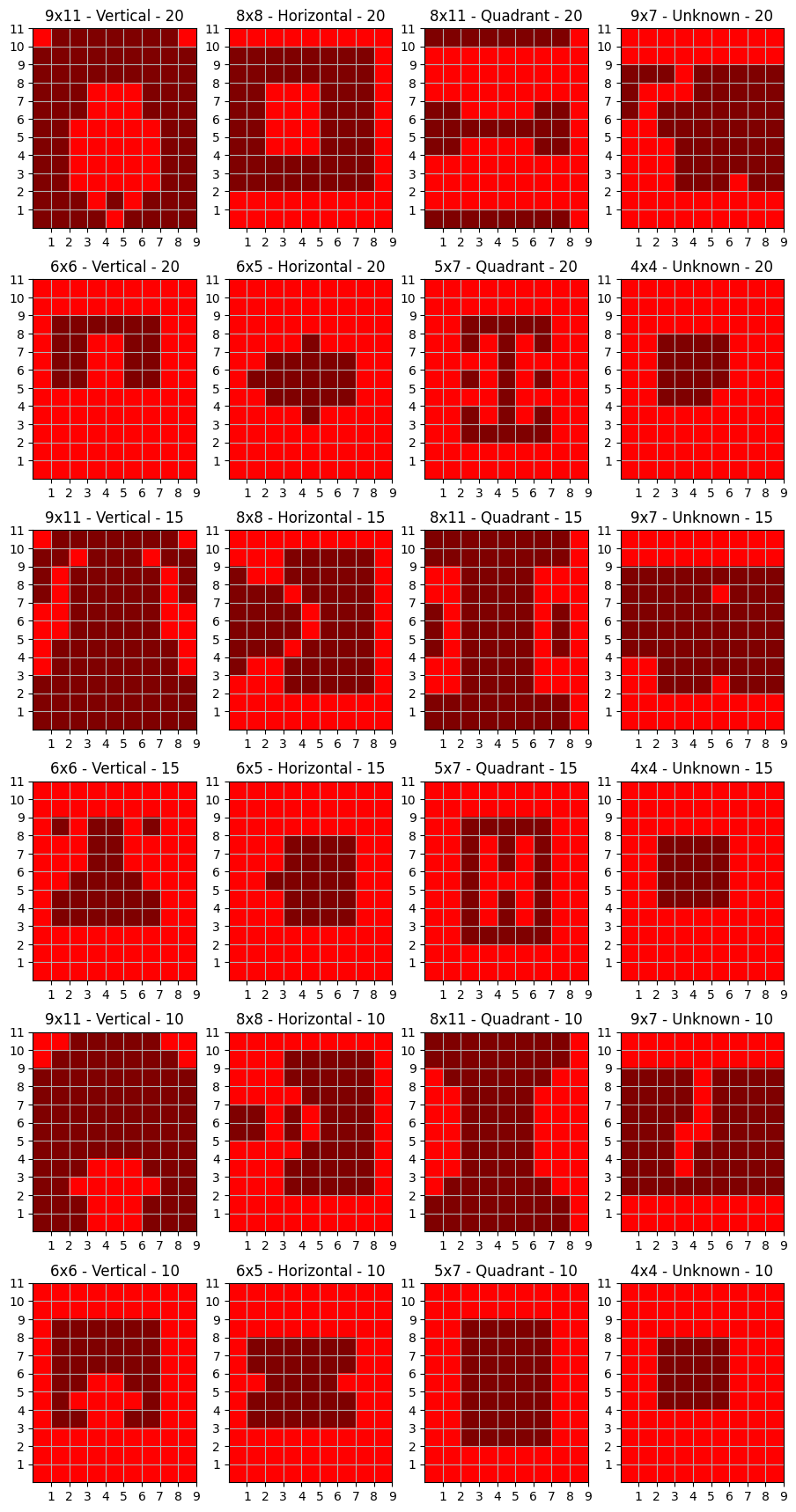}
    \caption{Additional \textit{Avalon} generated levels. We show examples using all types of symmetries (columns), $3$ levels of difficulty ($2$ rows per difficulty) and $8$ different sizes (repeated every $2$ rows).}
    \label{fig:avalon_additional_examples}
\end{figure}

\noindent The second comparison focuses on evaluating how well the network follows the style of the dataset. Given the lack of a clear pattern in the main dataset, we trained models using the stylized dataset, which contains very discernible patterns and is easier to evaluate visually. We evaluate these \textit{Stylized} models quantitatively and qualitatively and draw parallelisms between each pair of main and stylized versions.

From a qualitative point of view, the \textit{Stylized} generated levels contain more block-like structures than their counterparts, which show more vertically blocked configurations or single-blocked structures, as illustrated in Figures \ref{fig:inference_vanillastylized_example}-\ref{fig:inference_vanilla_example} and \ref{fig:inference_avalonstylized_example}-\ref{fig:inference_avalon_example} respectively. This follows the patterns contained in the training set (Figures \ref{fig:dataset_main_example} and \ref{fig:dataset_stylized_example}). However, the results of the \textit{Stylized} generators clearly show that the well-defined $2\times2$ BLOCK pattern is not correctly generated.

From a quantitative perspective, the Stylized models achieve a lower tile distribution accuracy than their counterparts: $65.12\%$ compared to $39.58\%$ for the \textit{Vanilla} and \textit{VanillaStylized} generator and $55.81\%$ compared to $35.42\%$ for the \textit{Avalon} and \textit{AvalonStylized} generator. This shows a discrepancy between the quantitative metric and the qualitative results. We hypothesize that the spread of the main dataset's distribution is larger, and thus, easier for the median of the inference distribution to fall within the $[Q_1,Q_3]$ range of the training examples. Moreover, not all the layout sizes are as well represented in the training set (see Figure \ref{fig:training_heatmap}), which makes the distribution comparison imprecise.

Our partial generation strategy allows us to generate symmetric levels with the \textit{Stylized} generators even though the dataset used to train them did not contain any such examples. However, it is possible that a GAN model under this conditional framework would capture better patterns while featuring similar behaviors.

A more extensive exemplification of different combinations of symmetries, sizes, and moves generated by the \textit{Avalon} model can be found in Figure \ref{fig:avalon_additional_examples}.


\section{Conclusion and Future Work}
\label{sec:conclusion}


\noindent This paper presents Avalon, an auto-validation level generation framework that enhances generative models through conditioning using gameplay statistics during training. This method learns from both the style of previous examples and level-validating data generated automatically through bots.

We have applied our method to the generation of simplified match-3 layouts using a cVAE conditioned on game mechanic features, in this case the median number of moves to solve the level as a proxy for difficulty, and relevant visual features, like size and symmetry. Our novel partial generation approach effectively solves the symmetry generation problem, and conditioning on the median number of moves allows for more fine-grained controllability at inference time. Our results show that conditioning the cVAE on the difficulty achieves a percentage of $51.39\%$ valid levels. This is an improvement over an equivalent model without this condition ($43.75\%$) and the dataset baseline ($46.15\%$).

However, our results also suggest that adding a difficulty condition can decrease the performance of other requirements measured by our quantitative metrics, like size or tile distribution accuracy. This could be accentuated by sampling conditional features with contradictory effects 
and the lack of good data. As noticeably highlighted by Figure \ref{fig:training_heatmap} the training set is extremely sparse with respect to the condition parameters, and especially lacking in the small-sized range of valid levels.
The qualitative evaluation of the model trained on the stylized dataset shows that, even though high-level patterns in the dataset are captured, the model has a harder time with detailed low-level patterns.

The strategy for selecting a checkpoint does not currently account for the difficulty, since in our case, this is a very costly process. In cases where this metric can be easily computed in the learning framework, its addition to the group of metrics being averaged would help create a model that performs better in that area. Additionally, a user study could be carried out to analyze the effect of different weights when averaging the evaluation metrics. Models with different characteristics could be used in different design scenarios. To avoid retraining models, checkpoints can be saved during training according to particular weighting strategies.

The comparison between the models with and without difficulty conditioning assumes equivalent architectures. However, generating levels according to the number of moves necessary for solving them is more complex. The \textit{Avalon} method might benefit from a higher capacity architecture to be able to model these complex relationships and more training data, to deal better with the harder learning task. A better approach for creating synthetic data that covers the solution space with respect to the conditioning parameters would doubtlessly help with the data problem, too.
A complementary exploration of the proposed framework using alternative generative models could improve the ability to capture style patterns more accurately than the cVAE we implemented. 

Further, using RL agents trained to play match-3 games, e.g. \citet{karimi2022candyrl}, could provide improved validation capabilities for the difficulty conditioning. Heuristic bots do not necessarily capture the dynamic gameplay strategies that could emerge through RL. These strategies may be closer to those of human players, aligning better the number of moves needed to solve a level in the desired use case scenario, with human players. For games in production, player or tester data could also be tested in this framework.

The next step for the match-3 generation would be to solve the generation of complete levels. To do so, the representation needs to account for complex and stacked objects. This could be done with a multi-layer representation, similar to that of \citet{candy_crush} or \citet{lilys_garden}, which is already compatible with our method. One interesting approach would be to use a chain of networks to generate subsequent layers conditioned on the output of the previous network or input from the designer. This format would support both mixed initiative generation and a fully automated solution. The step-by-step generation would affect the final difficulty of the level depending on how the layers interact with each other. For example, it would allow designers to start from an easy layout and add elements that would increase or decrease the difficulty of the level, e.g. using blockers or power-ups.

Finally, it would undoubtedly be helpful for the community if this approach were to be tested on different game genres. Here, we present some examples. For games widely studied in the literature, like Super Mario Bros. or Zelda, the difficulty proxy can be represented by a combination of completion time and number of enemies. For other genres, like racing games the difficulty could be defined as the time difference with the best-performing rival. In `capture the flag' modes in other genres, the win rate balance between the teams might be appropriate. The definition of the problem will depend on the needs of level designers.

\section{Acknowledgments}
\noindent We would like to express our gratitude to Tracktwenty for providing data, engineering help, and expertise around level design in match-3 games. We would also like to thank Judith Bütepage for proofreading the paper and for helpful discussions during the development of the method.



\bibliographystyle{IEEEtranN}
{\small \bibliography{refs}}

\end{document}